\documentclass[letterpaper, 10 pt, conference]{ieeeconf}  

\IEEEoverridecommandlockouts                              
\usepackage[utf8]{inputenc}
\usepackage[T1]{fontenc}
\usepackage{graphicx}
\usepackage{float} 
\usepackage{ragged2e}
\usepackage{booktabs}
\usepackage{bbding}
\usepackage{multirow}
\usepackage{makecell}
\usepackage{amsmath,amssymb,amsfonts}
\usepackage[table,xcdraw]{xcolor}
\usepackage{cite}
\usepackage[colorlinks,linkcolor=blue]{hyperref}
\usepackage{cleveref}
\usepackage{tabularx}
\usepackage[caption=true,font=normalsize,labelfont=sf,textfont=sf]{subfig}
\usepackage{textcomp}
\usepackage{stfloats}
\usepackage{tabularx}
\usepackage[linesnumbered,ruled,vlined]{algorithm2e}
\usepackage{adjustbox} 
\usepackage{pifont}

\hypersetup{
    colorlinks=true, 
    linkcolor=blue,  
    filecolor=blue,  
    urlcolor=black,   
    citecolor=blue,  
    pdfborder={0 0 0} 
}

\Crefname{figure}{Fig.}{Figs.}
\Crefname{equation}{Eq.}{Eqs.}

\def\ie{\emph{i.e.}}

\newcommand{\verticaltext}[3][0pt]{%
  \rotatebox[origin=l]{90}{\vspace{#1} #2 #3}%
}

\newtheorem{theorem}{Theorem}
\newcommand{\xmark}{\ding{55}}

\title{\bf FedDSR: Federated Deep Supervision and Regularization \\ Towards Autonomous Driving}

\author{~~~Wei-Bin Kou$^{1,2,3}$, Guangxu Zhu$^{3}$, Bingyang Cheng$^{1}$, Chen Zhang$^{1}$, Yik-Chung Wu$^{1,*}$, and Jianping Wang$^{2,*}$
\thanks{$^{*}$Corresponding author: Yik-Chung Wu (ycwu@eee.hku.hk) and Jianping Wang (jianwang@cityu.edu.hk).}
\thanks{$^{1}$Department of Electrical and Electronic Engineering, The University of Hong Kong, Hong Kong 999077, China.}
\thanks{$^{2}$Department of Computer Science, City University of Hong Kong, Hong Kong 999077, China.}
\thanks{$^{3}$Shenzhen International Center For Industrial And Applied Mathematics, Shenzhen Research Institute of Big Data, Shenzhen, China.}
}

\begin{document}

\maketitle
\thispagestyle{empty}
\pagestyle{empty}

\begin{abstract}
Federated Learning (FL) enables collaborative training of autonomous driving (AD) models across distributed vehicles while preserving data privacy. However, FL encounters critical challenges such as poor generalization and slow convergence due to non-independent and identically distributed (non-IID) data from diverse driving environments. To overcome these obstacles, we introduce \underline{Fed}erated \underline{D}eep \underline{S}upervision and \underline{R}egularization (FedDSR), a paradigm that incorporates multi-access intermediate layer supervision and regularization within federated AD system. Specifically, FedDSR comprises following integral strategies: (I) to select multiple intermediate layers based on predefined architecture-agnostic standards. (II) to compute mutual information (MI) and negative entropy (NE) on those selected layers to serve as intermediate loss and regularizer. These terms are integrated into the output-layer loss to form a unified optimization objective, enabling comprehensive optimization across the network hierarchy. (III) to aggregate models from vehicles trained based on aforementioned rules of (I) and (II) to generate the global model on central server. By guiding and penalizing the learning of feature representations at intermediate stages, FedDSR enhances the model generalization and accelerates model convergence for federated AD. We then take the semantic segmentation task as an example to assess FedDSR and apply FedDSR to multiple model architectures and FL algorithms. Extensive experiments demonstrate that FedDSR achieves up to 8.93\% improvement in mIoU and 28.57\% reduction in training rounds, compared to other FL baselines, making it highly suitable for practical deployment in federated AD ecosystems.
\end{abstract}

\section{INTRODUCTION}
The advent of autonomous driving (AD) technologies has revolutionized transportation, enabling vehicles to perceive their surroundings, make decisions, and navigate safely without human intervention \cite{kou2024adverse,10416354,9981567,10160999,9811702,10049523}. Central to these capabilities are deep learning models that interpret complex sensor inputs, including images from cameras, point clouds from LiDAR, and radar signals, for essential tasks like road scene semantic segmentation, object detection, lane tracking, etc. However, training such models traditionally requires centralized data aggregation, which poses significant privacy risks as vehicles generate sensitive location-based data. Federated Learning (FL) \cite{https://doi.org/10.48550/arxiv.1602.05629,kou2024fedrc,zhang2022lsfl} emerges as a viable solution, allowing distributed vehicles to collaboratively learn a shared global model by aggregating vehicles' updates at a central server, while keeping raw data private.

Despite its promise, FL in AD encounters an inherent challenge that undermine its wide application in AD. Specifically, non-independent and identically distributed (non-IID) data distribution is prevalent \cite{10611202}. For example, a vehicle in a city might encounter abundant pedestrian and traffic light data, while one in a rural area deals primarily with open roads and natural obstacles, leading to distributed skewed datasets. This heterogeneity in data distribution results in a global model with poor generalization and slow convergence when deployed fleet-wide. 

\begin{figure*}[tp]
\centering
\includegraphics[width=\linewidth]{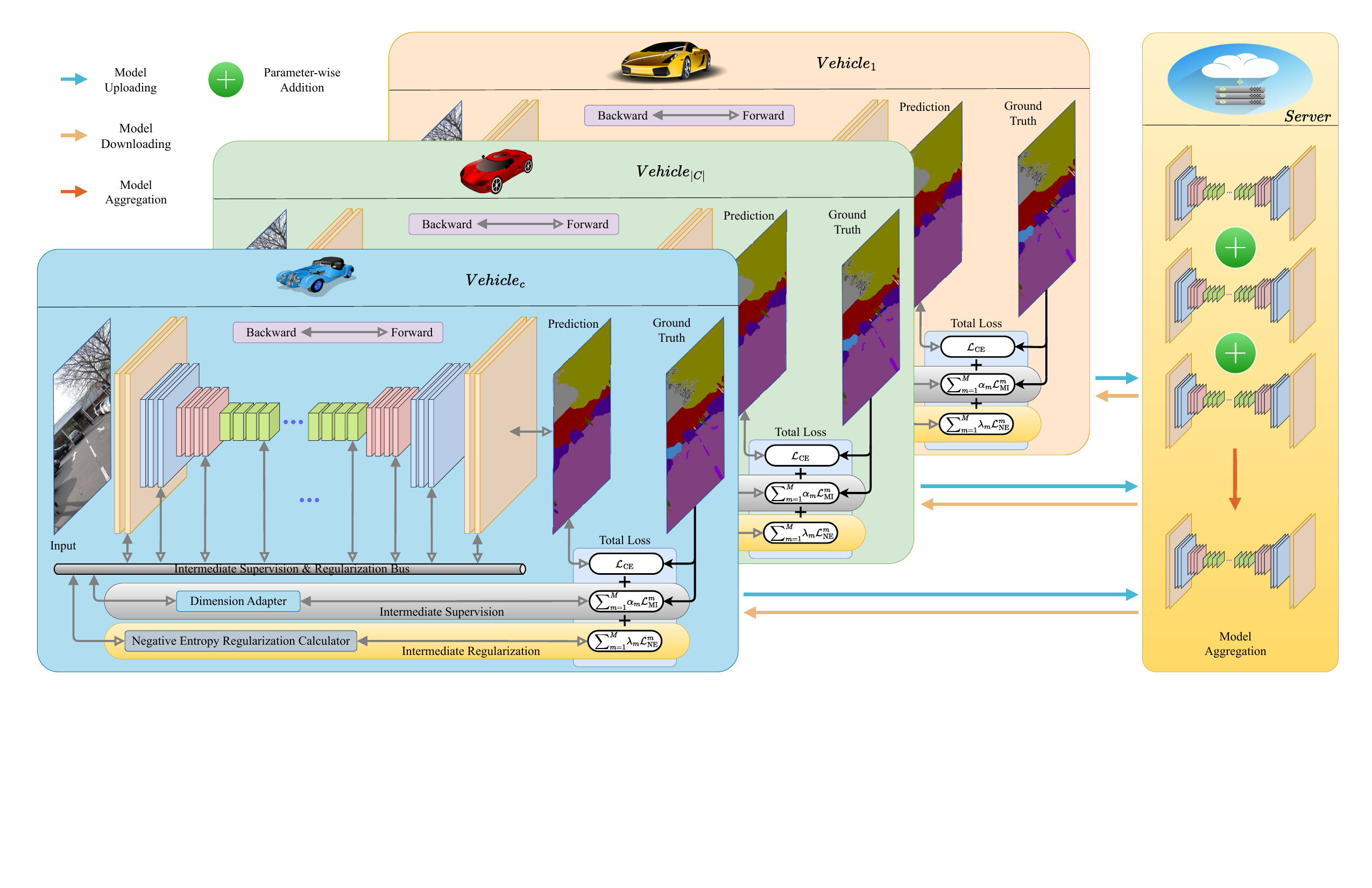}
\vspace{-3.4cm}
\caption{Overview of the proposed FedDSR.}
\label{Fig:FedDSR_overview}
\vspace{-0.5cm}
\end{figure*}

To mitigate this FL intrinsic challenge in AD, we propose \underline{Fed}erated \underline{D}eep \underline{S}upervision and \underline{R}egularization (FedDSR) scheme from optimization perspective. The proposed FedDSR derives from the fundamental limitation in deep model training, where intermediate layers often receive insufficient supervision and regularization, leading to suboptimal feature extraction and overconfident predictions across vehicles under non-IID scenarios. To enhance the intermediate supervision and regularization in FL, the proposed FedDSR comprises following integral strategies: (I) Intermediate Point Selection: we select multiple intermediate layers based on predefined architecture-agnostic criteria to ensure flexibility across various model designs. For example, we can choose layers at key transition points in the network, such as layers between ResNet stages or transformer layers. (II) Intermediate Losses and Regularizers at Selected Intermediate Points: on the one hand, mutual information (MI) serving as loss is applied to intermediate layers to guide them to focus on learning diverse and hierarchical representations; on the other hand, negative entropy (NE) on the latent features of each intermediate point serving as a regularizer helps to improve generalization by penalizing overconfident hidden feature predictions. (III) Vehicles' Local Training: between two adjacent FL aggregations, each vehicle trains its local model using aforementioned strategies (I) and (II) based on local private dataset for multiple iterations. Those intermediate losses and regularizers are combined into the output-layer training loss to form a unified optimization objective, enabling comprehensive optimization across the network hierarchy. (IV) Federated Aggregation until Convergence: the central server aggregates the received models from all involved vehicles for multiple rounds until the global aggregated model converges. Within each round, the central server exchanges models with all participated vehicles to learn knowledge from all vehicles while without sharing private data. The synergy of abovementioned strategies (I) to (IV) helps to improve FL model generalization and accelerate FL model convergence in non-IID AD scenarios thanks to its innovative intermediate supervision and regularization.

In addition, we carry out a theoretical convergence analysis for FedDSR, providing insights into how the proposed FedDSR impacts the convergence of FL model training. The proposed FedDSR is illustrated in \Cref{Fig:FedDSR_overview}. We evaluate FedDSR based on semantic segmentation task, and apply FedDSR to multiple model architectures and compare FedDSR with multiple existing FL algorithms.

Main contributions of this work are highlighted as follows:
\begin{itemize}
    \item To improve FL model generalization and accelerate FL model convergence in non-IID AD scenarios, we propose FedDSR, a FL training scheme that integrates both supervision and regularization on multiple architecture-agnostic intermediate layers. 
    \item We additionally conduct theoretical convergence analysis for FedDSR, which suggests that FedDSR holds $\mathcal{O}(1/\sqrt{T})$ convergence rate that matches standard SGD optimization, proving FedDSR does not harm asymptotic convergence. 
    \item We take AD semantic segmentation task as an example to assess the proposed FedDSR, and apply it to multiple model architectures and FL algorithms. Extensive experiments show that FedDSR achieves an 8.93\% improvement in mIoU and a 28.57\% reduction in training rounds compared to other FL baselines. In addition, we conduct ablation studies to explore how the number of intermediate points, the distance between adjacent intermediate points, and positions of the intermediate points affect the model performance.
\end{itemize}

\section{Related Work}
\label{related_work}

\subsection{Federated Autonomous Driving (FedAD) System}
FL \cite{10342134,feddrive2022,11127787,kou2025pfedlvm} emerges as an innovative solution to bolster the generalization of deep learning model. In the context of AD, FL aggregates model parameters from multiple vehicles to harness distributed data without centralization. This approach enables AD vehicles to share knowledge of new data samples or edge cases with a centralized server and other vehicles while maintaining data privacy. Foundational frameworks like FedAvg \cite{https://doi.org/10.48550/arxiv.1602.05629} facilitate this by having vehicles perform local stochastic gradient descent (SGD) on their data and upload model updates for server-side aggregating. However, FedAvg falters in common non-IID AD scenarios, where data imbalances cause performance drops and convergence slow-downs. To counter these, FedProx \cite{li2020federated} incorporates a proximal term in the local objective to penalize deviations from the global model; Scaffold \cite{karimireddy2020scaffold} introduces control variates to correct for client drift, improving variance reduction; FedRC \cite{10342134} accelerates model convergence by integrating Gaussian distribution to measure data heterogeneity; FedEMA \cite{kou2025fedema} improves generalization by incorporating historical model fitting capabilities; personalized FL \cite{kou2025pfedlvm} strategies adapt global models to local data via fine-tuning, useful for vehicle-specific customizations; Client clustering \cite{kou2025fast} groups similar data distributions to reduce non-IID effects. 
Despite these advancements, limitations persist in handling architectural diversity and non-IID data in real-time vehicular FL, where poor generalization and slow convergence can impede FL's wide application in AD. To widen its application, we propose FedDSR to incorporate intermediate supervision and regularization to improve model generalization and accelerate model convergence under non-IID AD scenarios. 

\subsection{Deep Supervision and Regularization}
Deep supervision techniques have been pivotal in centralized deep learning for AD tasks, with early methods applying auxiliary classifiers to intermediate layers to combat gradient vanishing issues~\cite{hochreiter2001gradient}, thereby enhancing training stability in deep networks~\cite{lee2015deeply,zhang2022contrastive,li2022comprehensive}. For example, GoogleNet~\cite{szegedy2015going} incorporates two additional supervision layers at intermediate stages; DSN~\cite{wang2015training} introduces auxiliary supervision branches at specific intermediate layers; PSPNet~\cite{Zhao_2017_CVPR} use auxiliary losses for pyramid pooling modules by incorporating an auxiliary classifier to calculate pixel-wise cross-entropy between predictions and ground truth; BiSeNet~\cite{yu2018bisenet} balances spatial and contextual paths with supervised branches to ensure harmony between detail and global context; Gated-SCNN~\cite{9009833} introduces shape-based intermediate losses to enhance shape-aware features; and ICNet~\cite{Zhao_2018_ECCV} adds auxiliary loss branches to low-resolution intermediate predictions in a cascaded framework. Complementary regularization approaches address overconfidence by penalizing peaked distributions in activations to promote uncertainty awareness vital for uncertain driving conditions like fog or low light. While deep supervision and regularization have broad applications, these methods are predominantly centralized, and there exists a notable gap in applying these deep supervision and regularization techniques to FL scenarios. This paper presents FedDSR to mitigate this gap.

\section{Methodology}
\label{methodology}
In this section, we firstly elaborate the proposed FedDSR in \Cref{FedDSR_formulation}. We subsequently conduct convergence analysis for the proposed FedDSR in \Cref{FedDSR_convergence}. 

\subsection{FedDSR Formulation and Optimization} \label{FedDSR_formulation}
The proposed FedDSR embeds supervision and regularization at multiple intermediate points of the deep learning model, fostering resilient feature hierarchies that generalize across non-IID distributions. FedDSR comprises one central server and multiple vehicles, where the central server coordinates training among all vehicles, each with local dataset reflecting respective unique driving environments. The key notations of FedDSR formulation are summarized in \Cref{tab:FedDSR_notations}. 

\subsubsection{Intermediate Supervision and Regularization Point Selection}
FedDSR firstly selects some intermediate supervision and regularization points based on some predefined model architecture-independent rules. Generally, layers at key transition points are preferred, as these points often undertake significant changes in feature representation. Examples include but not limit to:  
\begin{itemize}
    \item \textbf{Before or after downsampling} (e.g., pooling layers) to capture changes in spatial size and feature granularity.  
    \item \textbf{Between major blocks} (e.g., ResNet stages or transformer layers) to leverage the differences in feature abstraction between hierarchical stages.  
    \item \textbf{At bottleneck layers}, where the feature dimensions are compressed, highlighting critical information.  
    \item \textbf{Before or after attention mechanisms} to capture how signal is distributed or aggregated across feature maps.
    \item \textbf{At feature fusion points} in multi-branch architectures, to capture the integration of diverse feature streams.  
\end{itemize}  

These transition points provide a comprehensive view of how features evolve throughout the network, enabling more effective learning of latent feature representations.

\begin{table}[t]
    \centering
    \setlength{\tabcolsep}{5.0pt}
    \caption{Key Notations of FedDSR Formulation}
    \begin{tabularx}{\linewidth}{ll}
    \hline
        \textbf{Symbols} & \textbf{Definitions} \\ \hline
        $N$ & Number of vehicles in Federated AD systems \\ 
        $\mathcal{D}_n$ & Local dataset on vehicle $n$ \\ 
        $|\mathcal{D}_n|$ & Number of samples on vehicle $n$ \\ 
        $\mathcal{D} = \bigcup_{n} \mathcal{D}_n$ & Global data distribution (implicit; data remain local) \\ 
        $(x_i, y_i)$ & Input image and pixel-wise semantic label pair \\ 
        $\theta$ & Global model parameters \\ 
        $\theta_n^t$ & Vehicle $n$'s local model parameters in round $t$ \\ 
        $M$ & Number of intermediate supervision/regularization points \\ 
        $G_m$ & Intermediate supervision and regularization point $m$ \\ 
        $z^m$ & Latent feature maps at $G_m$ \\ 
        $C_m$ & Number of channels used by NE at $G_m$ \\ 
        $p_k(x_i; \theta)$ & Output-layer softmax probability for class $k$ \\ 
        $q_k^m(\cdot; \phi_m^n)$ & Dimension adapter for $G_m$ with parameters $\phi_m^n$ \\ 
        $\mathcal{L}_{\text{CE}}^n$ & Cross-entropy loss on vehicle $n$ \\ 
        $\mathcal{L}_{\text{MI}}^{m,n}$ & Mutual information loss for point $G_m$ on vehicle $n$ \\ 
        $\mathcal{L}_{\text{NE}}^{m,n}$ & Negative entropy regularization for point $G_m$ on vehicle $n$ \\ 
        $\alpha_m, \lambda_m$ & Loss weights for $G_m$ \\ 
        $\eta$ & Learning rate \\ 
        $E$ & Local epochs per round \\ 
        $B$ & Batch size \\ 
        $T$ & Total FL training rounds \\ 
        $S^t$ & All involved vehicles in round $t$ \\ 
        $w_n$ & Federated aggregation weight for vehicle $n$ \\ \hline
    \end{tabularx}
\label{tab:FedDSR_notations}
\end{table}

\subsubsection{Definitions of Intermediate Losses and Regularizers}
FedDSR considers a central server and $N$ vehicles, where vehicle $n$ holds a private dataset $\mathcal{D}_n$ and trains a local model $\theta_n$ with multiple intermediate supervision and regularization points $\{G_1, \dots, G_M\}$. 
Specifically, for vehicle $n$, the components of local optimization objective are as follows:
\begin{itemize}
    \item \textbf{Output-layer cross-entropy (CE):}
    \begin{equation}
    \mathcal{L}_{\text{CE}}^n = -\frac{1}{|\mathcal{D}_n|} \sum_{i=1}^{|\mathcal{D}_n|} \sum_{p \in \Omega} \sum_{k=1}^K y_{i,p,k} \log p_k(x_{i,p}; \theta_n),
    \label{eq:L_CE}
    \end{equation}
    where $\Omega$ is the pixel set of images and $K$ is the number of semantic classes.
    
    \item \textbf{Intermediate mutual information (MI) loss at $G_m$:}
    \begin{equation}
    \!\mathcal{L}_{\text{MI}}^{m,n} \!=\! -\frac{1}{|\mathcal{D}_n|}\! \sum_{i=1}^{|\mathcal{D}_n|} \sum_{p \in \Omega} \sum_{k=1}^K y_{i,p,k} \log q_k^m(z_{i,p}^m; \phi_m^n),
    \label{eq:L_MI}
    \end{equation}
    where $z_{i,p}^m = \theta_n^{G_m}(x_i)$ is the latent feature at point $G_m$, $\phi_m^n$ is resolution adapter to calculate MI.

    \item \textbf{Intermediate negative entropy (NE) regularization at $G_m$:}
    \begin{equation}
    \!\!\!\mathcal{L}_{\text{NE}}^{m,n}\! =\! \frac{1}{|\mathcal{D}_n|} \!\sum_{i=1}^{|\mathcal{D}_n|} \!\sum_{p \in \Omega} \!\sum_{c=1}^{C_m} \!p_c^m\!(z_{i,p}^m; \!\theta_n) \!\log\! p_c^m\!(z_{i,p}^m; \!\theta_n),
    \label{eq:L_NE}
    \end{equation}
    where $p_c^m(z_{i,p}^m; \theta_n)$ is the softmax probability at pixel $p$ for channel $c$ at the intermediate point $G_m$.
\end{itemize}

To summarize, the total local objective for client $n$ is
\begin{equation}
\mathcal{L}^n = \mathcal{L}_{\text{CE}}^n + \sum_{m=1}^M \left( \alpha_m \mathcal{L}_{\text{MI}}^{m,n} + \lambda_m \mathcal{L}_{\text{NE}}^{m,n} \right),
\label{eq:local_obj}
\end{equation}
where $\alpha_m$ and $\lambda_m$ are the weights for intermediate loss and regularizer at point $G_m$.

\subsubsection{Optimization of FedDSR}
The global optimization objective of FedDSR is
\begin{equation}
\min_{\theta, \Phi} \sum_{n=1}^N w_n \mathcal{L}^n,
\end{equation}
where $\Phi = \{\phi_m\}_{m=1}^M$ is the set of all involved adapters, and $w_n = |\mathcal{D}_n| / \sum_{n=1}^N |\mathcal{D}_n|$ are the aggregation weights.

The training process of FedDSR follows below steps:
\begin{itemize}
    \item \textbf{Server broadcast:} In round $t$, the server sends global parameters $\theta^{t-1}$ to all vehicles in $S^t$.
    \item \textbf{Client-side local training:} Each vehicle $n \in S^t$:
    \begin{enumerate}
        \item Initializes: $\theta_n^t \leftarrow \theta^{t-1}$;
        \item Trains locally for $E$ epochs with batch size $B$, minimizing $\mathcal{L}^n$;
        \item Updates parameters using gradient descent:
        \begin{equation}
        \theta_n^t \leftarrow \theta_n^{t} - \eta \nabla_{\theta_n^t} \mathcal{L}^n;
        \end{equation}
        \item Sends updated parameters $\theta_n^t$ to the server.
    \end{enumerate}
    \item \textbf{Server aggregation:} The server aggregates parameters of all received models:
    \begin{equation}
    \theta^{t+1} = \sum\nolimits_{n \in S^t} w_n \theta_n^t.
    \end{equation}
\end{itemize}

\begin{algorithm}[tp]
\caption{FedDSR}
\label{alg:FedDSR}
\SetKwInOut{KwIn}{Input}
\SetKwInOut{KwOut}{Output}
\KwIn{Vehicles $\{1,\dots,N\}$ with local datasets $\{\mathcal{D}_n\}_{n=1}^N$; global model $\theta$ with intermediate points $\{G_1,\dots,G_M\}$; learning rate $\eta$; local epochs $E$; batch size $B$; total rounds $T$}
\KwOut{Trained global model ${\theta^*}$}

\textbf{Initialization:} Initialize $\theta^0$, $\Phi^0$, $\{\alpha_m,\lambda_m\}_{m=1}^M$\;

\For{round $t=0$ \KwTo $T-1$}{
    Sample participating vehicles $S^t \subseteq \{1,\dots,N\}$\;
    Broadcast $(\theta^t, \Phi^t)$ to all $n \in S^t$\;

    \tcp{Parallel training for vehicles}
    \ForPar{each vehicle $n \in S^t$}{
        $\theta_n \leftarrow \theta^t$\;

        \For{local epoch $e=1$ \KwTo $E$}{
            \For{$(x_i, y_i) \in\ $B$\ \in \mathcal{D}_n$}{
                \tcp{Forward Pass}
                $\{z_{i}^1,\dots,z_{i}^M\} \leftarrow \theta_n(x_i)$\;
                $p_k(x_i; \theta_n) \leftarrow \mathrm{Softmax}(\theta_n(x_i))$\;

                $\mathcal{L}_{\text{CE}}^n \leftarrow $ \Cref{eq:L_CE}\;
                \For{$m=1$ \KwTo $M$}{
                    $\mathcal{L}_{\text{MI}}^{m,n} \leftarrow$ \Cref{eq:L_MI}, $\mathcal{L}_{\text{NE}}^{m,n} \leftarrow$ \Cref{eq:L_NE}\;
                }
                $\mathcal{L}_{\text{T}}^n \leftarrow$ \Cref{eq:local_obj}\;

                \tcp{Backward Pass}
                Compute $\nabla_{\theta_n}\mathcal{L}^n$, $\nabla_{\phi_{m,n}}\mathcal{L}^n$ for all $m$\;
                \For{$m=1$ \KwTo $M$}{
                    $\phi_{m,n} \leftarrow \phi_{m,n} - \eta \nabla_{\phi_{m,n}} \mathcal{L}^n$\;
                }
                $\theta_n \leftarrow \theta_n - \eta \nabla_{\theta_n} \mathcal{L}^n$\;
            }
        }
        Send $(\theta_n, |\mathcal{D}_n|)$ to server\;
    }

    \tcp{Server-side aggregation}
    $w_n \leftarrow |\mathcal{D}_n|/\sum_{n \in S^t} |\mathcal{D}_n|$ for $n \in S^t$\;
    $\theta^{t+1} \leftarrow \sum_{n \in S^t} w_n \theta_n$\;
}
\Return $\theta^* \leftarrow \theta^{T}$
\end{algorithm}

In conclusion, the training process of FedDSR is outlined in Algorithm \ref{alg:FedDSR}.

\subsection{Convergence Analysis of FedDSR} \label{FedDSR_convergence}

The global optimization objective of FedDSR is
\begin{equation}
\mathcal{L}(\theta) = \sum\nolimits_{n=1}^N w_n \, \mathcal{L}^{n}(\theta).
\end{equation}
To clearly conduct convergence analysis of FedDSR, some vehicle-side assumptions are made. Specifically, for each component $s \in \{\text{CE}, \{_{\text{MI}}^{m,}\}, \{_{\text{NE}}^{m,}\}\}$ and each vehicle $n \in \{1,\dots,N\}$, the local loss $\mathcal{L}_s^{n}$ satisfies:
\begin{itemize}
    \item \textbf{L-Smoothness:} There exists $\mathcal{L}_s^{n} > 0$ such that $\forall \theta,\theta'$, $\|\nabla \mathcal{L}_s^{n}(\theta) - \nabla \mathcal{L}_s^{n}(\theta')\| \le \mathcal{L}_s^{n} \|\theta - \theta'\|$.
    \item \textbf{Bounded Stochastic Gradients:} There exists $G_s^{n} > 0$ such that $\forall \theta$ and minibatch $\mathcal{B}$ sampled from $\mathcal{D}_n$, $\mathbb{E}[\|\nabla \mathcal{L}_s^{n}(\theta;\mathcal{B})\|^2] \le (G_s^{n})^2$.
    \item \textbf{Bounded Variance:} There exists $(\sigma_s^{n})^2 > 0$ such that $\forall \theta$ and $\mathcal{B}$, $\mathbb{E}[\|\nabla \mathcal{L}_s^{n}(\theta;\mathcal{B}) - \nabla \mathcal{L}_s^{n}(\theta)\|^2] \le (\sigma_s^{n})^2$.
\end{itemize}

On top of these local assumptions, the global smoothness constant is defined as $\mathcal{L}_s = \sum_{n=1}^N w_n \mathcal{L}_s^{n}$, and let $\mathcal{L}_{\text{max}} = \max\{\mathcal{L}_{\text{CE}}, \alpha_m \mathcal{L}_{\text{MI}}^m, \lambda_m \mathcal{L}_{\text{NE}}^m\}$, where $\mathcal{L}_{\text{CE}} = \sum_{n=1}^N w_n \mathcal{L}_{\text{CE}}^{n}$, $\mathcal{L}_{\text{MI}}^m = \sum_{n=1}^N w_n \mathcal{L}_{\text{MI}}^{m,n}$, and $\mathcal{L}_{\text{NE}}^m = \sum_{n=1}^N w_n \mathcal{L}_{\text{NE}}^{m,n}$. Similarly, we define
\begin{align}
\!\!G_{\text{T}}^2 &\!=\!\!\!\sum_{n=1}^N \!w_n\!(\!(G_{\text{CE}}^{n})^2 \!\!+\!\!\sum_{m=1}^M \!\!(\alpha_m^2 (G_{\text{MI}}^{m,n})^2 \!\!+\!\!\lambda_m^2 (G_{\text{NE}}^{m,n})^2\!)\!), \\
\!\!\sigma_{\text{T}}^2 &\!=\!\!\!\sum_{n=1}^N \!w_n\! (\!(\sigma_{\text{CE}}^{n})^2 \!\!+ \!\!\sum_{m=1}^M\!\! (\alpha_m^2 (\sigma_{\text{MI}}^{m,n})^2 \!\!+\!\! \lambda_m^2 (\sigma_{\text{NE}}^{m,n})^2\!)\!).
\end{align}

We consider FL aggregation with partial participation $S^t \subseteq \{1,\dots,N\}$ in round $t$, $E$ local SGD epochs with step-size $\eta_t$, and aggregation weights $w_n = \frac{|\mathcal{D}_n|}{\sum_{n \in S^t} |\mathcal{D}_n|}$ for $n \in S^t$. Let $\theta_t$ represents global model in round $t$. Based on these assumptions, we can conclude below Theorem 1 about the convergence rate of the proposed FedDSR.

\begin{theorem}
\label{thm:federated_convergence}
Suppose the above assumptions hold and each vehicle performs $E$ local SGD steps with step-size $\eta_t = \frac{\eta}{\sqrt{T}}$ at round $t$, and the server aggregates models by weighted summation. Then after $T$ federated aggregation rounds, we have
\begin{align}
\!\!\frac{1}{T} \!\sum_{t=1}^T \mathbb{E}\| \nabla \mathcal{L}(\theta_t) \|^2 \!
\le \!\!\underbrace{\frac{2\Delta}{\eta \sqrt{T}}}_{\text{Initial Gap}}\!\!
+\! \underbrace{\frac{\mathcal{L}_{\text{max}} \eta}{\sqrt{T}} ( G_{\text{T}}^2\!\! +\!\! \sigma_{\text{T}}^2 \!\!+\!\! \Gamma_{\text{drift}} )}_{\text{Variance and Drift Terms}},
\end{align}
where $\Delta = \mathcal{L}(\theta_0) - \mathcal{L}^*$, $\theta_0$ is the initial model parameters, $\mathcal{L}^*$ is the theoretical optimal value for $\mathcal{L}(\theta)$, and $\Gamma_{\text{drift}}$ captures the additional error due to client drift from multiple local steps and partial participation. A common bound for $\Gamma_{\text{drift}}$ under $E$ local steps is
\begin{equation}
\!\!\Gamma_{\text{drift}} \;\le\; c \, E^2 \, (\underbrace{\mathbb{E}\|\nabla \mathcal{L}(\theta_t)\|^2}_{\text{Global Gradient Magnitude}} \;\!\!+\!\!\; \underbrace{\mathcal{H}}_{\text{Data Heterogeneity}}),
\end{equation}
for some constant $c>0$, and
\begin{equation}
\mathcal{H} \;=\; \sum\nolimits_{n=1}^N w_n \, \|\nabla \mathcal{L}^{n}(\theta_t) - \nabla \mathcal{L}(\theta_t)\|^2,
\end{equation}
which measures vehicle heterogeneity (non-IIDness). Consequently, FedDSR achieves the standard $\mathcal{O}(1/\sqrt{T})$ rate for non-convex objectives, up to additional drift terms controlled by $E$ and heterogeneity.
\end{theorem}

From Theorem 1, we can conclude following insights: (I) With fixed $E$ and bounded heterogeneity, the bound preserves the $\mathcal{O}(1/\sqrt{T})$ convergence rate as in conventional non-convex SGD. (II) The convergence bound scales with the number of intermediate points $M$ via $G_{\text{T}}^2$ and $\sigma_{\text{T}}^2$, thus, choosing $M=\mathcal{O}(\log D)$ for model depth $D$ controls this growth. (III) Client drift increases with $E$, therefore, the trade-off between communication and computation can be tuned by selecting moderate $E$.

Owing to the space limit, we just offer the primary proof sketch of Theorem 1 in the \textbf{\textit{Appendix}}.

\section{Experiments}
\label{experiments}
In this section, we take semantic segmentation task as an example to evaluate FedDSR. All evaluations are based on widely recognized datasets, model architectures, FL algorithms, and metrics. We firstly introduce the experimental setup in \Cref{sec:exp_setup}. Subsequently, we reveal the experimental results and conduct analyses in \Cref{sec:main_results}. Finally, we conduct ablation studies in \Cref{sec:ablation}.

\subsection{Datasets, Evaluation Metrics and Implementation} \label{sec:exp_setup}
\subsubsection{Datasets}
The Cityscapes dataset \cite{Cordts2016Cityscapes} consists of 2,975 training images and 500 validation images, each annotated with masks. This dataset encompasses 19 semantic classes, such as vehicles and pedestrians. The CamVid dataset \cite{brostow2008segmentation} comprises a total of 701 images across 11 semantic classes. For our experiments, we randomly selected 600 samples for training and used the remaining 101 samples as a test dataset. The SynthiaSF dataset \cite{ros2016synthia} offers a collection of synthetic, yet photorealistic images that emulate urban scenarios. It provides pixel-level annotations for 23 semantic classes, with 1,596 images designated for training and 628 for testing.

\subsubsection{Evaluation Metrics}
We assess the proposed FedDSR on semantic segmentation task using four commonly used metrics: mIoU, mPrecision (mPre for short), mRecall (mRec for short), and mF1.

\begin{table*}[tp]
\centering
\setlength{\tabcolsep}{5.45pt}
\caption{Quantitative performance comparison of FedDSR against FedAvg for multiple model architectures}
\begin{tabularx}{\linewidth}{c|c|c|cccc|cccc|cccc}
\hline
\multirow{2}{*}{Models} &\multirow{2}{*}{Backbone}    & \multirow{2}{*}{FL Algorithm} & \multicolumn{4}{c|}{CamVid Dataset (\%)}                        & \multicolumn{4}{c|}{Cityscapes Dataset (\%)}                           & \multicolumn{4}{c}{SynthiaSF Dataset (\%)}                        \\ \cline{4-7} \cline{8-11} \cline{12-15} 
                            &            &                       &mIoU   &mF1    &mPre   &mRec   &mIoU   &mF1    &mPre   &mRec   &mIoU   &mF1    &mPre   &mRec   \\ \hline
\multirow{2}{*}{DeepLabv3+} &\multirow{2}{*}{ResNet50} & FedAvg  &72.78  &80.24  &81.33  &79.47  &47.91  &56.31  &59.99  &55.58  &26.65  &31.94  &40.30  &30.46  \\ 
                          &  & FedDSR                            &\textbf{73.24}  &\textbf{80.70}  &\textbf{81.60}  &\textbf{80.34}  &\textbf{50.49}  &\textbf{59.70}  &\textbf{65.00}  &\textbf{57.82}  &\textbf{29.03}  &\textbf{34.90}  &\textbf{41.12}  &\textbf{32.92}  \\ \hline
\multirow{2}{*}{SeaFormer} &\multirow{2}{*}{-}  & FedAvg         &47.08  &53.39  &53.86  &53.91  &27.32  &30.95  &32.78  &31.85  &16.29  &21.05  &\textbf{28.98}  &21.29  \\
                          &  & FedDSR                            &\textbf{51.54}  &\textbf{59.40}  &\textbf{60.07}  &\textbf{60.36}  &\textbf{29.77}  &\textbf{34.18}  &\textbf{33.84}  &\textbf{35.39}  &\textbf{18.43}  &\textbf{23.20}  &28.33  &\textbf{22.88}  \\ \hline
\multirow{2}{*}{TopFormer} &\multirow{2}{*}{-}  & FedAvg         &56.60  &63.21  &71.66  &63.26  &32.29  &37.23  &\textbf{36.84}  &38.35  &21.60  &26.88  &\textbf{32.22}  &\textbf{26.71}  \\
                         &   & FedDSR                            &\textbf{58.85}  &\textbf{66.20}  &\textbf{72.02}  &\textbf{65.20}  &\textbf{32.33}  &\textbf{37.24}  &36.75  &\textbf{38.43}  &\textbf{21.63}  &\textbf{26.98}  &31.80  &25.81  \\ \hline
\end{tabularx}
\label{tab:FedDSR_quantitative_comp}
\vspace{-0.5cm}
\end{table*}

\begin{figure}[tp]
\vspace{-0.1cm}
\centering
\subfloat[\footnotesize mIoU]{\includegraphics[width=0.5\linewidth, height=0.3\linewidth]{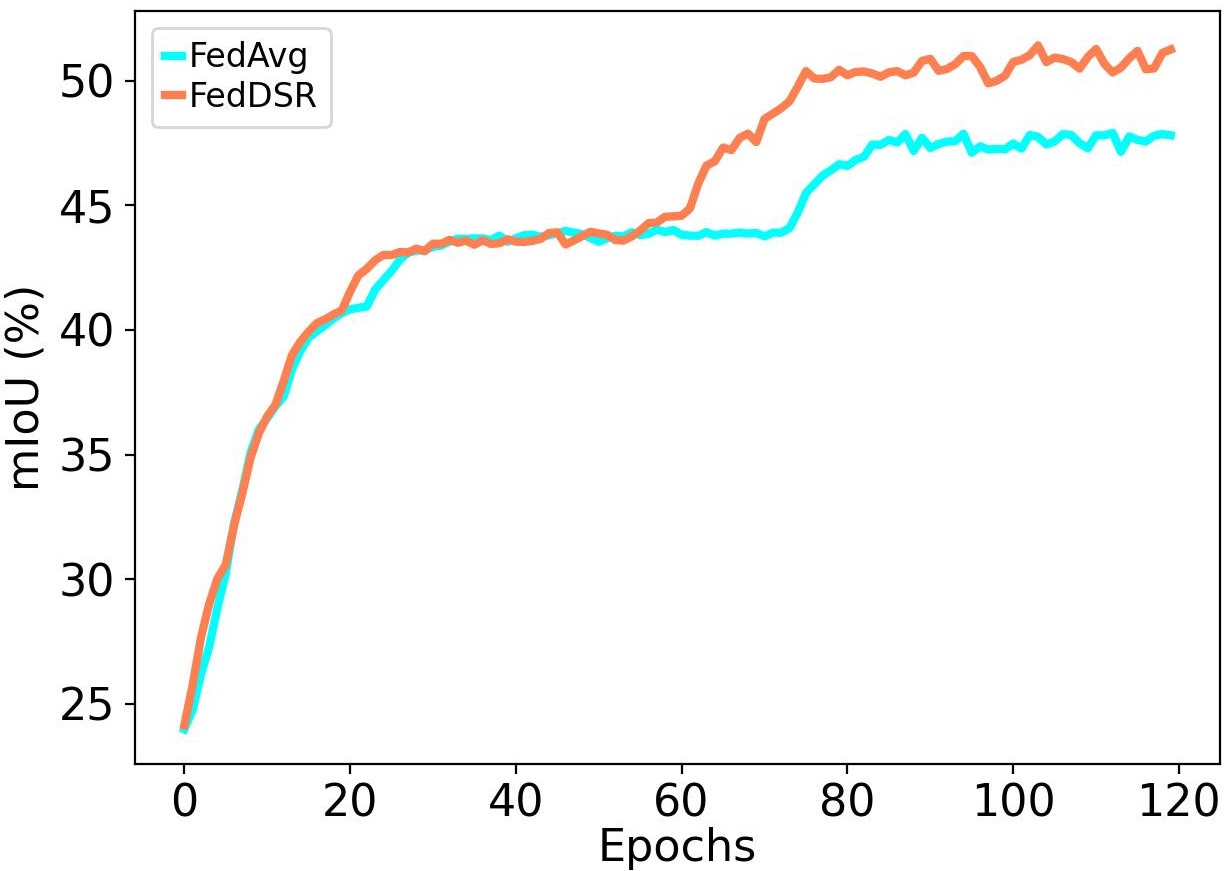}%
\label{Fig:city_mIoU}}
\subfloat[\footnotesize mF1]{\includegraphics[width=0.5\linewidth, height=0.3\linewidth]{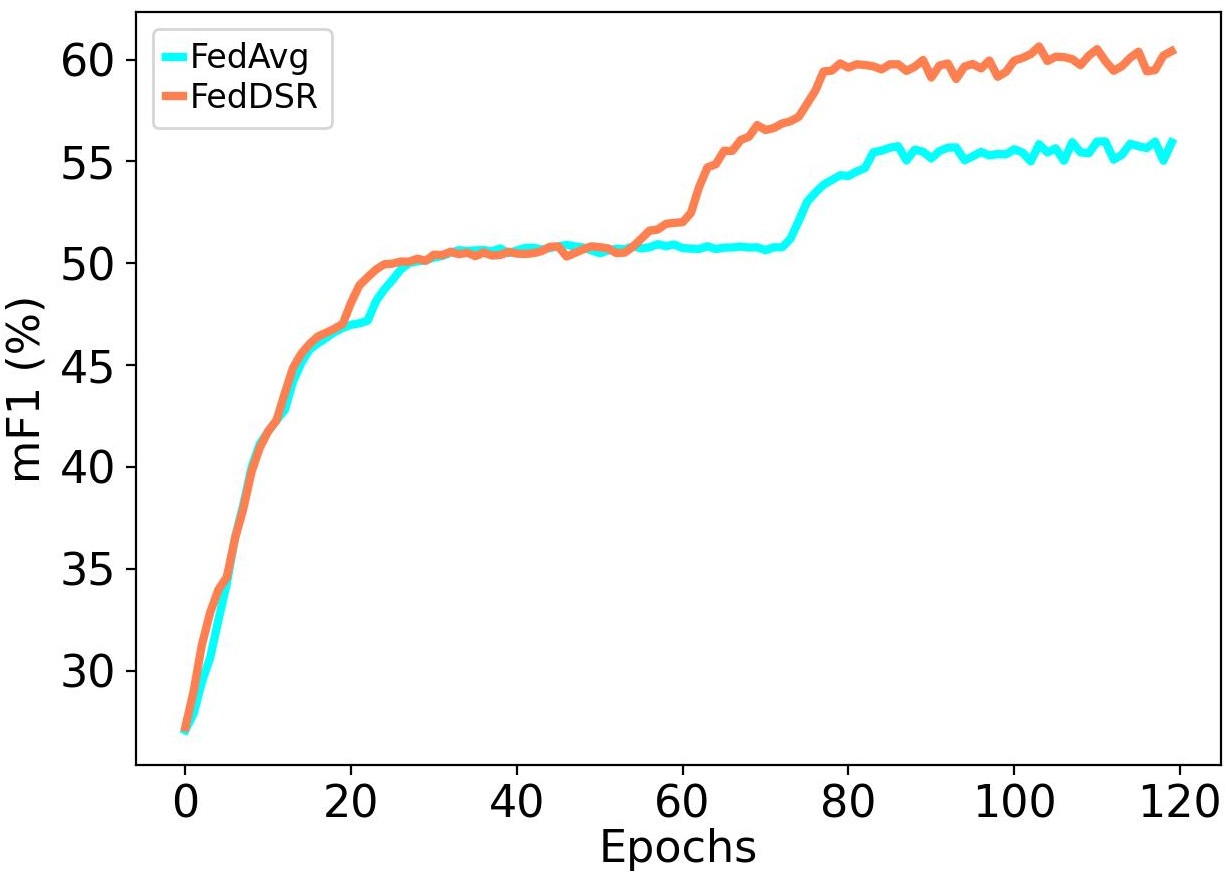}%
\label{Fig:city_mF1}}
\vspace{-0.2cm}
\caption{Convergence comparison of FedDSR against FedAvg.}
\label{Fig:convergece_comparison}
\vspace{-0.2cm}
\end{figure}

\begin{table}[tp]
\centering
\setlength{\tabcolsep}{7.1pt}
\caption{Quantitative performance comparison of FedDSR-enabled case against FedDSR-disabled case across multiple FL algorithms based on DeepLabv3+ architecture}
\begin{tabularx}{\linewidth}{c|c|cccc}
\hline
\multirow{2}{*}{FL Algorithm} & \multirow{2}{*}{FedDSR?} & \multicolumn{4}{c}{Cityscapes Dataset (\%)}                          \\ \cline{3-6} 
                            &                   &mIoU    &mF1     &mPre     &mRec     \\ \hline
\multirow{2}{*}{FedAvg} & \xmark                &47.91   &56.31   &59.99    &55.58    \\ 
                          & \checkmark          &\textbf{50.49}   &\textbf{59.70}   &\textbf{65.00}    &\textbf{57.82}    \\ \hline
\multirow{2}{*}{FedProx (0.005)} & \xmark       &41.41   &47.52   &47.69    &48.31    \\
                          & \checkmark          &\textbf{42.45}   &\textbf{48.53}   &\textbf{48.61}    &\textbf{49.36}    \\ \hline
\multirow{2}{*}{FedProx (0.01)} & \xmark        &31.74   &35.74   &35.82    &37.31    \\
                         & \checkmark           &\textbf{32.84}   &\textbf{36.82}   &\textbf{36.35}    &\textbf{38.38}    \\ \hline
\multirow{2}{*}{FedDyn (0.005)} & \xmark        &28.63   &31.95   &31.93    &33.95    \\
                         & \checkmark           &\textbf{29.75}   &\textbf{33.05}   &\textbf{32.31}    &\textbf{35.01}    \\ \hline
\multirow{2}{*}{FedDyn (0.01)} & \xmark         &25.19   &27.92   &26.63    &29.43    \\
                         & \checkmark           &\textbf{26.19}   &\textbf{28.92}   &\textbf{27.64}    &\textbf{30.43}    \\ \hline
\multirow{2}{*}{FedAvgM (0.7)} & \xmark         &51.49   &60.74   &65.37    &58.96    \\
                         & \checkmark           &\textbf{52.59}   &\textbf{61.84}   &\textbf{66.66}    &\textbf{59.93}    \\ \hline
\multirow{2}{*}{FedAvgM (0.9)} & \xmark         &51.65   &60.91   &65.55    &59.12    \\
                         & \checkmark           &\textbf{52.74}   &\textbf{61.93}   &\textbf{66.63}    &\textbf{60.39}    \\ \hline
\multirow{2}{*}{FedGau} & \xmark                &52.03   &62.23   &68.98    &59.36    \\
                         & \checkmark           &\textbf{54.40}   &\textbf{64.43}   &\textbf{71.94}    &\textbf{61.49}    \\ \hline
\multirow{2}{*}{FedIR} & \xmark                 &\textbf{25.83}   &\textbf{28.33}   &\textbf{27.91}    &\textbf{29.02}    \\
                         & \checkmark           &25.81   &28.32   &27.90    &28.98    \\ \hline
\multirow{2}{*}{MOON} & \xmark                  &51.48   &60.75   &65.37    &59.02    \\
                         & \checkmark           &\textbf{52.49}   &\textbf{61.67}   &\textbf{66.63}    &\textbf{59.70}    \\ \hline
\multirow{2}{*}{SCAFFOLD} & \xmark              &\textbf{24.19}   &\textbf{27.22}   &26.32    &\textbf{28.27}    \\
                         & \checkmark           &23.63   &26.79   &\textbf{26.38}    &27.49    \\ \hline
\multirow{2}{*}{BalanceFL} & \xmark             &\textbf{52.15}   &\textbf{61.36}   &\textbf{65.82}    &\textbf{59.53}    \\
                         & \checkmark           &51.48   &60.78   &65.54    &58.95    \\ \hline
\end{tabularx}
\label{tab:FedDSR_enhanced_FLs}
\vspace{-0.1cm}
\end{table}

\subsubsection{Implementation Details}
FedDSR and all adopted baselines are implemented using the Pytorch framework and are trained on two NVIDIA GeForce 4090 GPUs. For optimization, the Adam optimizer is chosen with Betas values of 0.9 and 0.999, and a weight decay of 1e-4.
Our experiments include a comparative analysis of the proposed FedDSR training method across three model architectures—DeepLabv3+ \cite{chen2018encoderdecoder}, TopFormer \cite{zhang2022topformer}, and SeaFormer \cite{wan2023seaformer}—on aforementioned datasets, namely Cityscapes, CamVid, and SynthiaSF. In addition, we also conduct comparison of FedDSR against other FL algorithms, including FedProx \cite{li2020federated}, FedDyn \cite{acar2021federated}, FedAvgM \cite{hsu2019measuring}, FedIR \cite{hsu2020federated}, MOON \cite{li2021model}, SCAFFOLD \cite{karimireddy2020scaffold}, FedAvg \cite{https://doi.org/10.48550/arxiv.1602.05629}, BalanceFL \cite{9825928}, and FedGau \cite{kou2025fast}.

\subsection{Main Results and Empirical Analyses} \label{sec:main_results}
This section presents experimental results on the FedDSR prediction performance, and conducts additional empirical analyses, from both quantitative and qualitative perspectives:

\subsubsection{Quantitative Evaluation}
\Cref{tab:FedDSR_quantitative_comp} presents quantitative performance of FedDSR against FedAvg for multiple model architectures, including DeepLabv3+, SeaFormer, and TopFormer. These evaluations use above discussed four metrics across CamVid, Cityscapes, SynthiaSF datasets. From \Cref{tab:FedDSR_quantitative_comp}, we can observe following patterns: (I) FedDSR outperforms FedAvg for almost all adopted model architectures by a large margin across CamVid, Cityscapes, and SynthiaSF datasets, which consolidates the superiority of the proposed FedDSR method. For example, for the experiment with DeepLabv3+ model architecture and SynthiaSF dataset, FedDSR improves FedAvg by (29.03-26.65) / 26.65 = 8.93\%, (34.90-31.94) / 31.94 = 9.27\%, (41.12-40.30) / 40.30 = 2.03\%, and (32.92-30.46) / 30.46 = 8.08\% in mIoU, mF1, mPre, and mRec, respectively. (II) The model architecture impacts the performance improvement of FedDSR over FedAvg. For example, for the experiments on SynthiaSF dataset, FedDSR can enhance the prediction performance for DeepLabv3+ and SeaFormer architectures while can not enhance that of TopFormer architecture, relative to FedAvg. (III) How much FedDSR enhances the performance relative to FedAvg somewhat depends on the complexity of experimental dataset. In general, the higher the complexity that the dataset has, the larger improvement that FedDSR achieves. For example, 
FedDSR achieves a notable improvement in prediction performance on the Cityscapes dataset with a 5.39\% increase in mIoU (\ie, (50.49 - 47.91) / 47.91=5.39\%), while delivering a smaller performance gain of (73.24 - 72.78) / 72.78=0.63\% in mIoU on the CamVid dataset, for the DeepLabv3+ architecture.

\begin{table*}[tp]
\centering
\renewcommand{\arraystretch}{0.24}
\addtolength{\tabcolsep}{-0.52pt}
\caption{Qualitative performance comparison of FedDSR-enabled setup against FedDSR-disabled setup}
\begin{tabularx}{\linewidth}{|l|lllll|}
\hline
\verticaltext[27.5pt]{\hspace{0.2cm}Raw RGBs} &
\includegraphics[width=0.188\linewidth, height=0.12\linewidth]{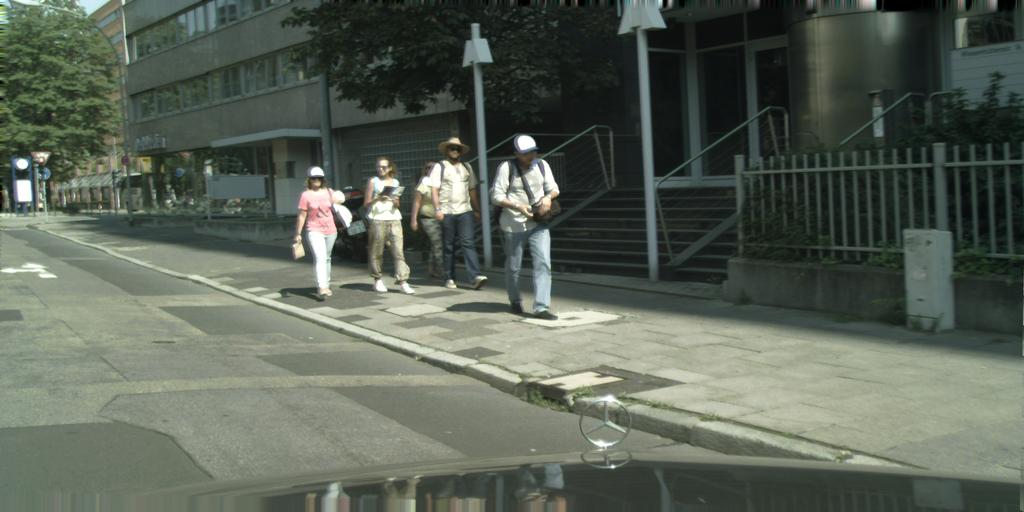} &\hspace{-0.47cm}
\includegraphics[width=0.188\linewidth, height=0.12\linewidth]{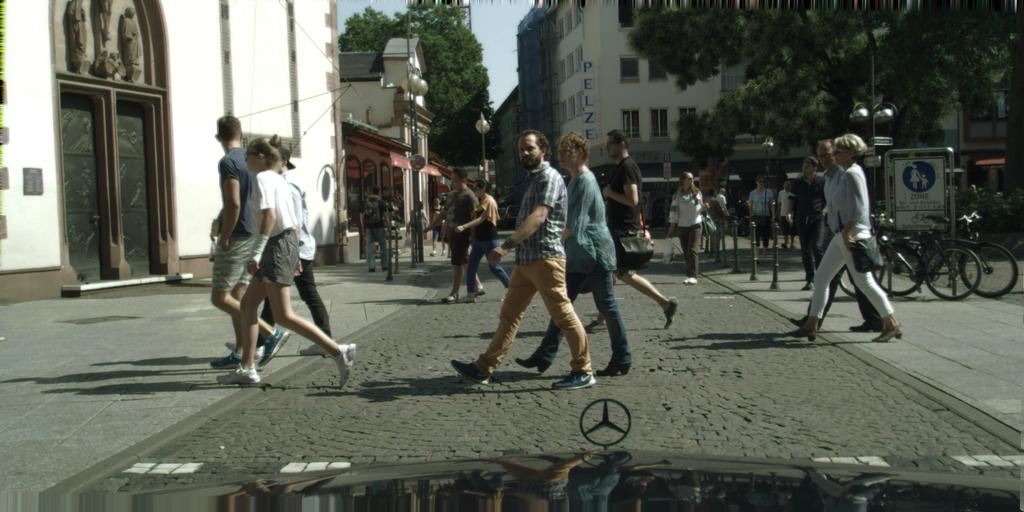} &\hspace{-0.47cm}
\includegraphics[width=0.188\linewidth, height=0.12\linewidth]{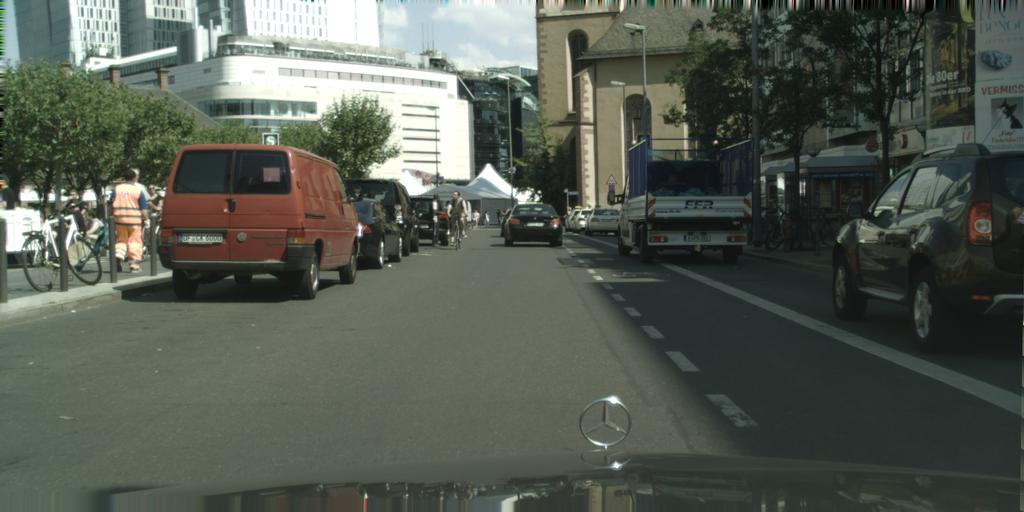} &\hspace{-0.47cm}
\includegraphics[width=0.188\linewidth, height=0.12\linewidth]{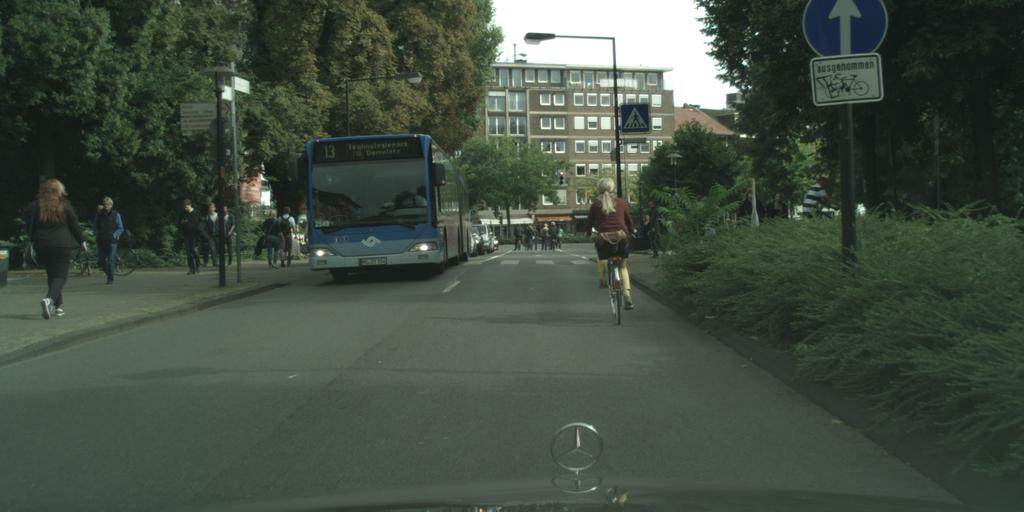} &\hspace{-0.47cm}
\includegraphics[width=0.188\linewidth, height=0.12\linewidth]{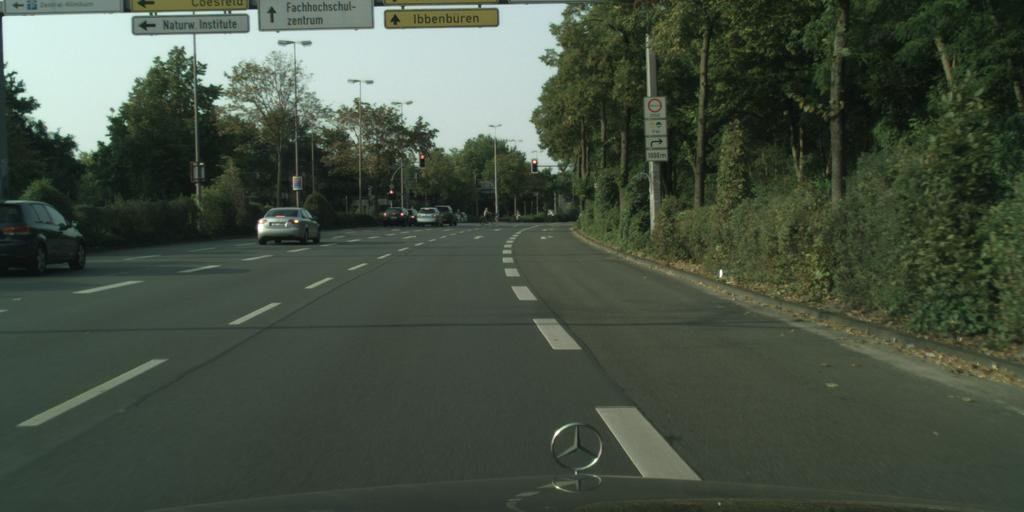}\\
\hline

\verticaltext[27.5pt]{\hspace{0.1cm}Ground Truth} &
\includegraphics[width=0.188\linewidth, height=0.12\linewidth]{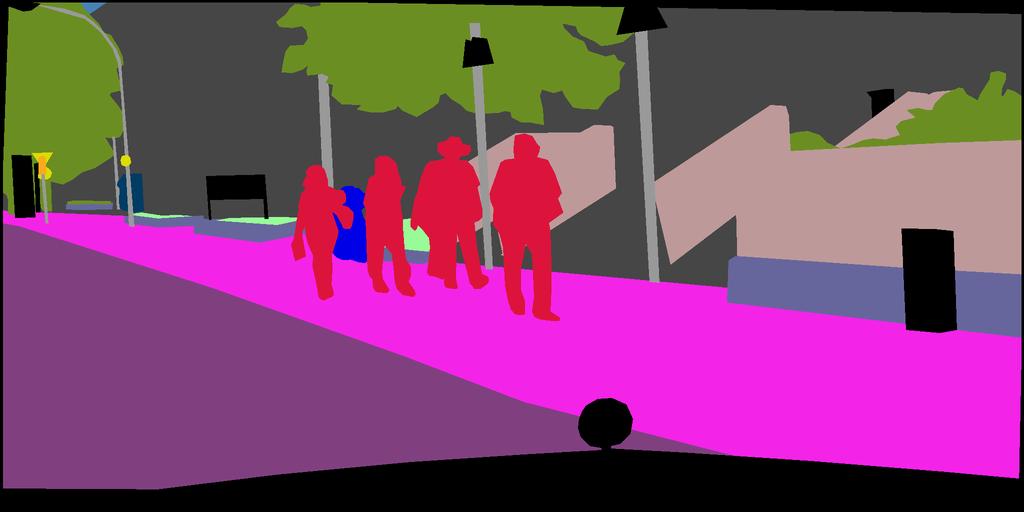} &\hspace{-0.47cm}
\includegraphics[width=0.188\linewidth, height=0.12\linewidth]{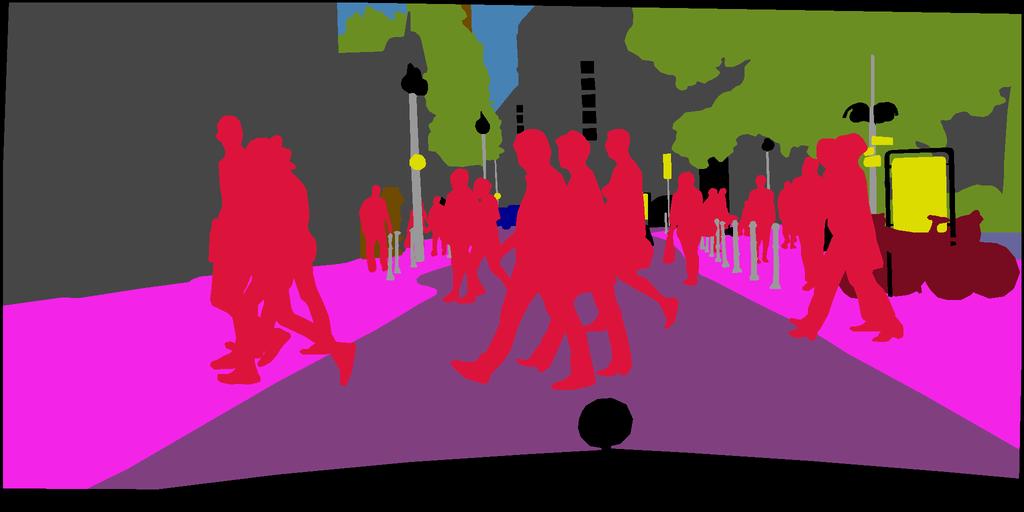} &\hspace{-0.47cm}
\includegraphics[width=0.188\linewidth, height=0.12\linewidth]{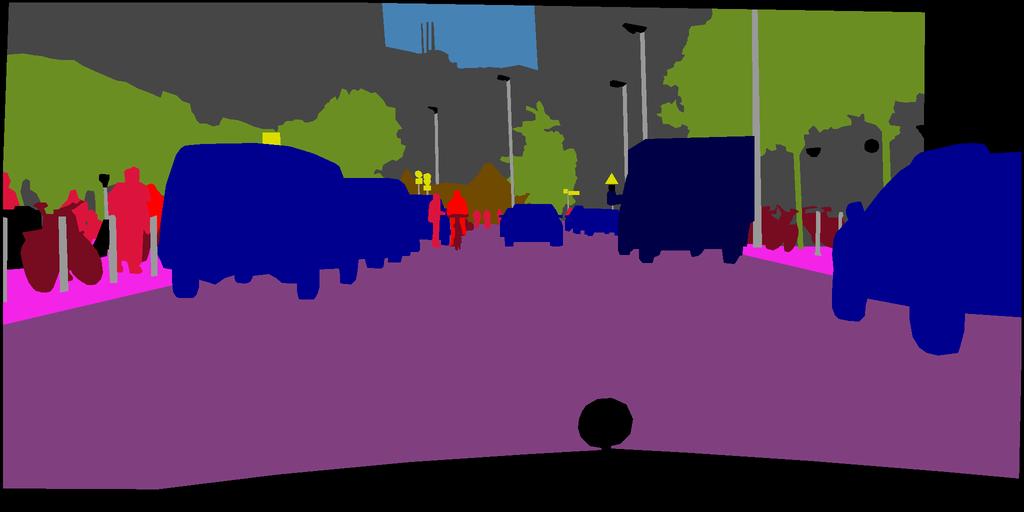} &\hspace{-0.47cm}
\includegraphics[width=0.188\linewidth, height=0.12\linewidth]{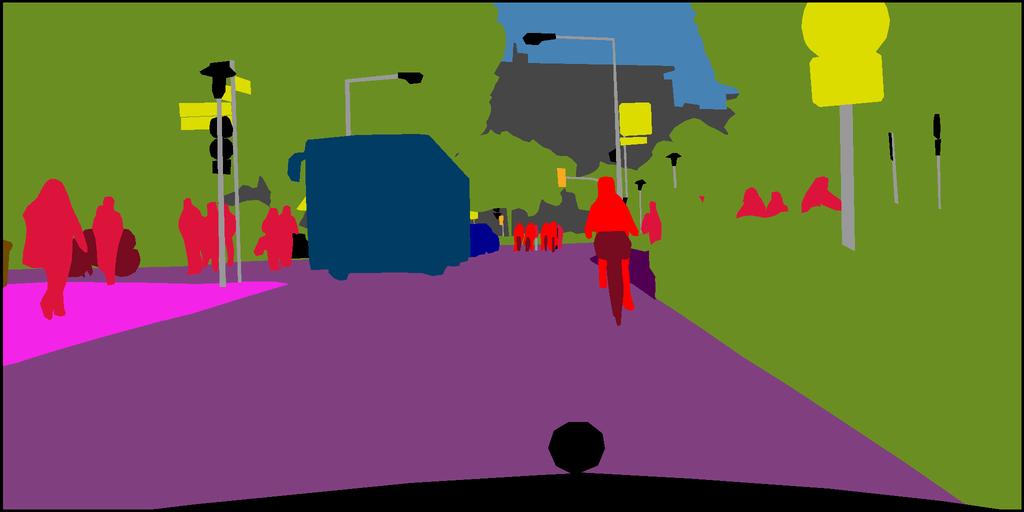} &\hspace{-0.47cm}
\includegraphics[width=0.188\linewidth, height=0.12\linewidth]{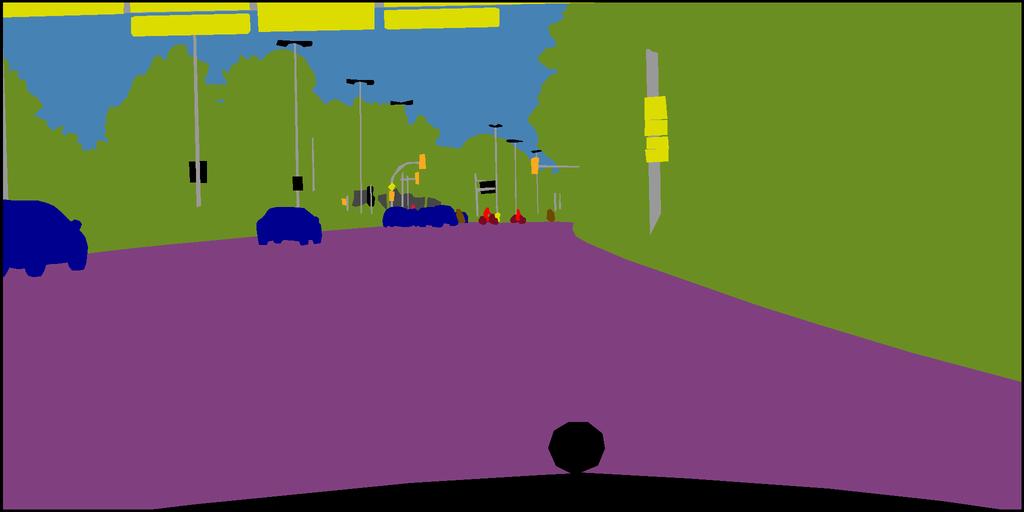}\\
\hline

\verticaltext[27.5pt]{\hspace{-0.15cm}FedDSR-Disabled} &
\includegraphics[width=0.188\linewidth, height=0.12\linewidth]{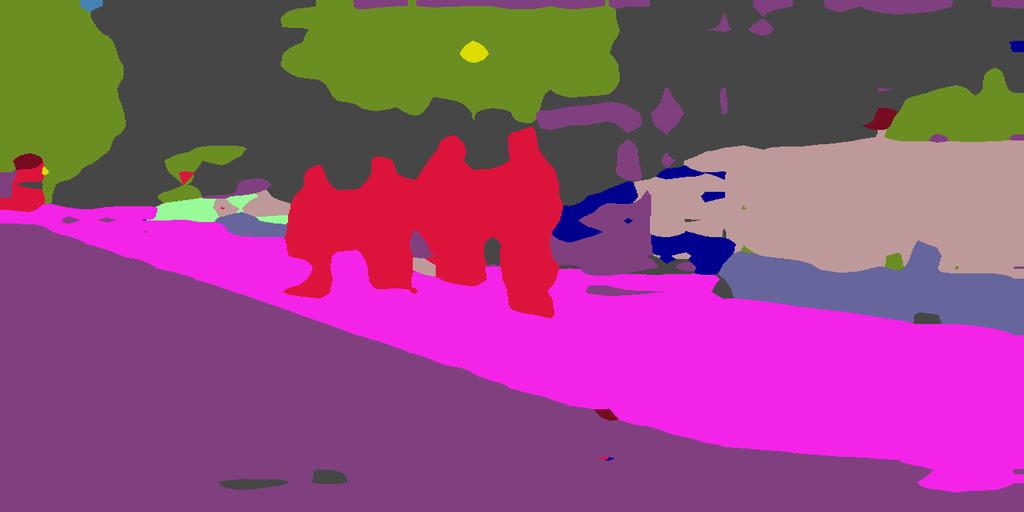} &\hspace{-0.47cm}
\includegraphics[width=0.188\linewidth, height=0.12\linewidth]{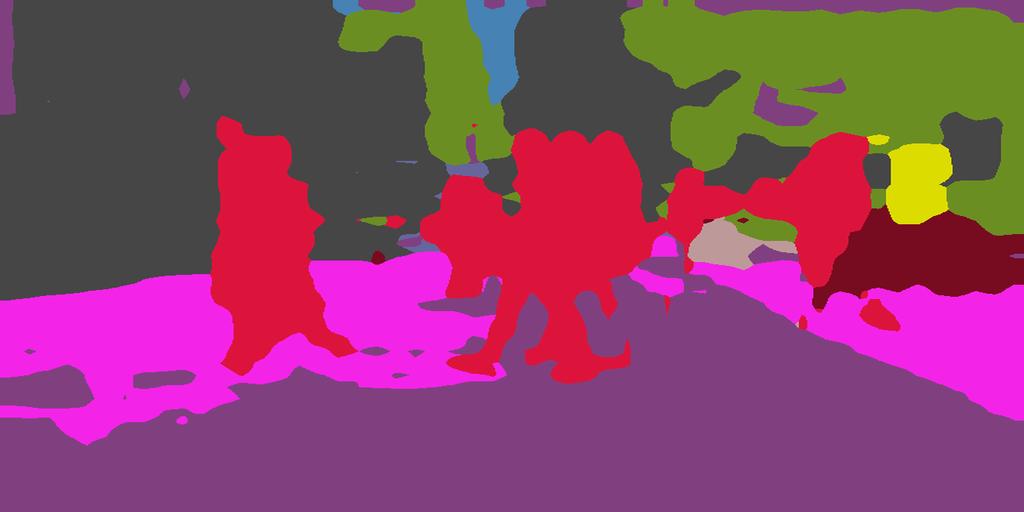} &\hspace{-0.47cm}
\includegraphics[width=0.188\linewidth, height=0.12\linewidth]{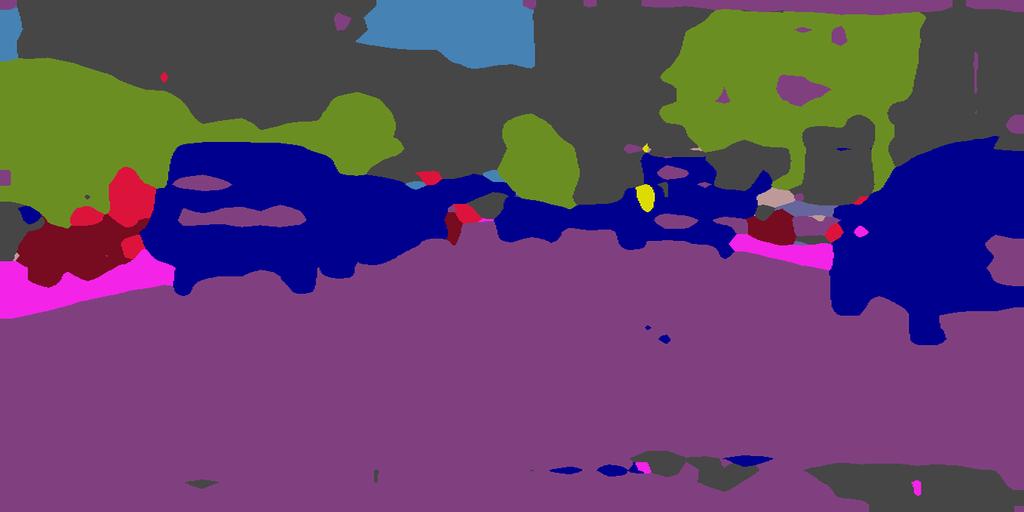} &\hspace{-0.47cm}
\includegraphics[width=0.188\linewidth, height=0.12\linewidth]{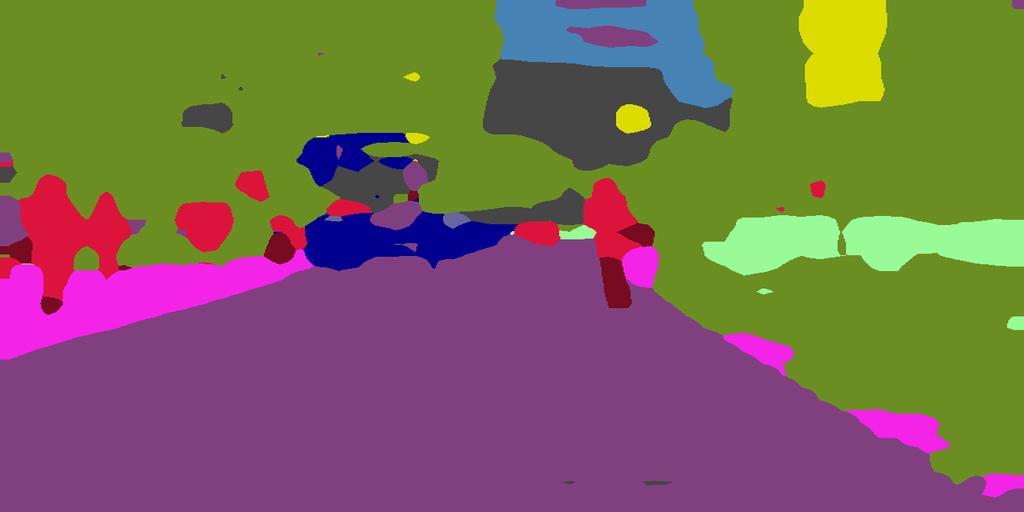} &\hspace{-0.47cm}
\includegraphics[width=0.188\linewidth, height=0.12\linewidth]{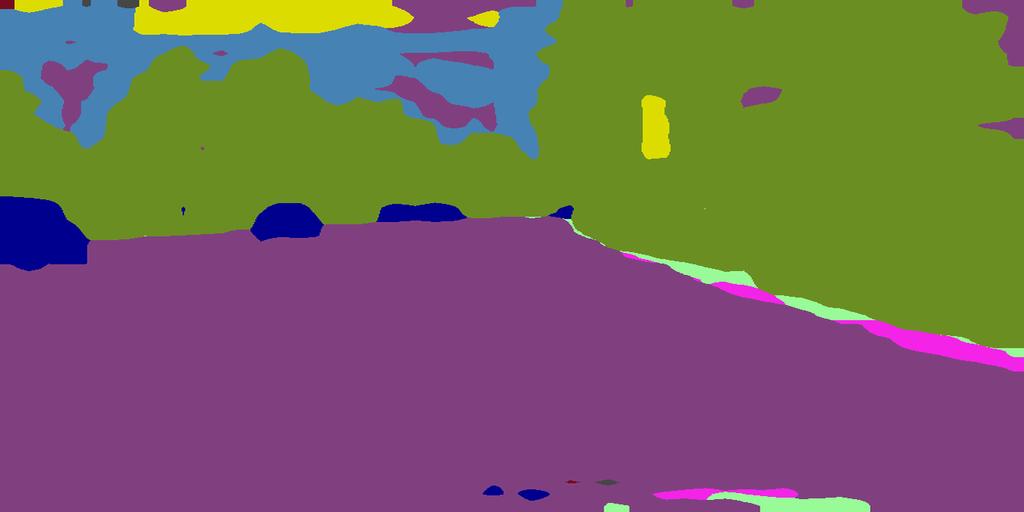}\\
\hline

\verticaltext[27.5pt]{\hspace{-0.18cm}\textbf{FedDSR-Enabled}} &
\includegraphics[width=0.188\linewidth, height=0.12\linewidth]{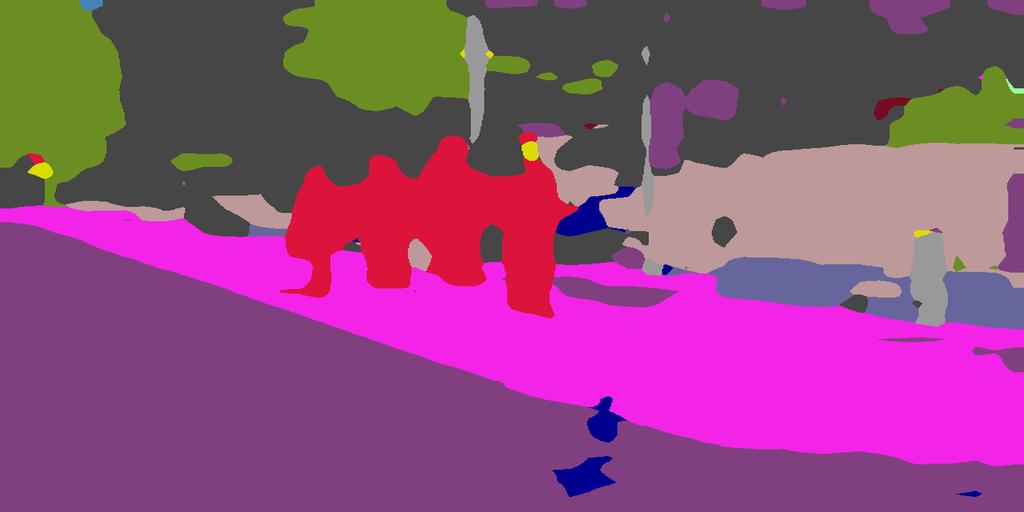} &\hspace{-0.47cm}
\includegraphics[width=0.188\linewidth, height=0.12\linewidth]{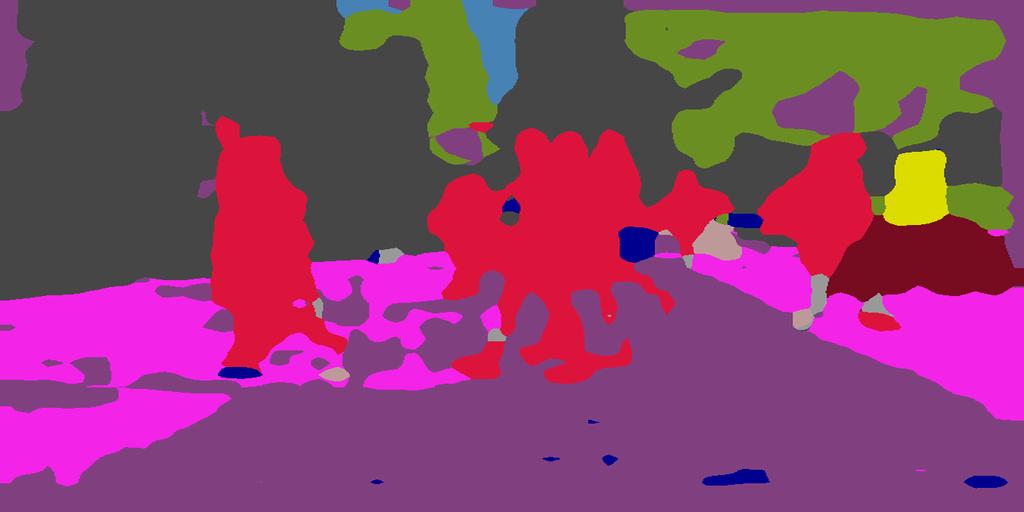} &\hspace{-0.47cm}
\includegraphics[width=0.188\linewidth, height=0.12\linewidth]{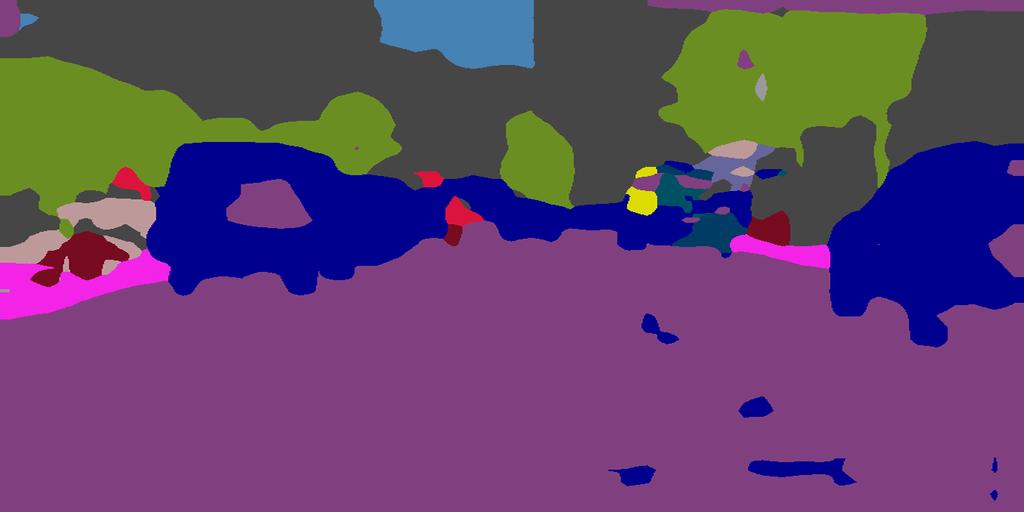} &\hspace{-0.47cm}
\includegraphics[width=0.188\linewidth, height=0.12\linewidth]{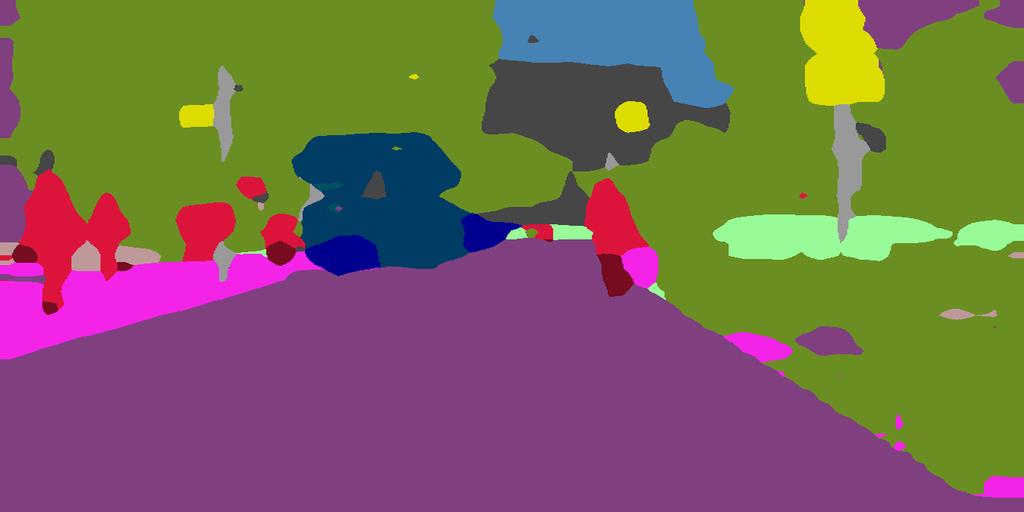} &\hspace{-0.47cm}
\includegraphics[width=0.188\linewidth, height=0.12\linewidth]{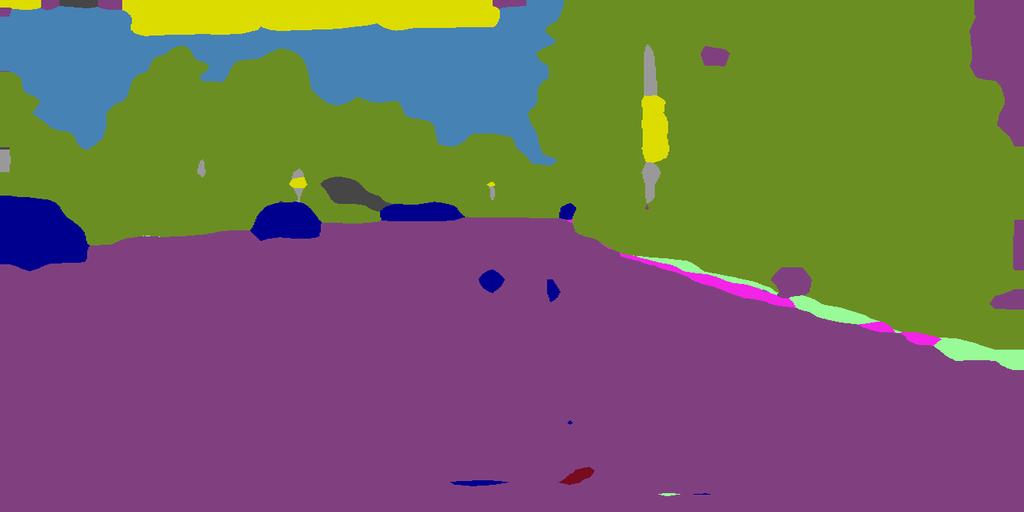}\\
\hline
\end{tabularx}
\label{tab:FedDSR_qualitive_comp}
\vspace{-0.6cm}
\end{table*}

\begin{table}[tp]
\centering
\setlength{\tabcolsep}{6.8pt}
\caption{Quantitative performance comparison of FedDSR setups with different number of intermediate points}
\begin{tabularx}{\linewidth}{c|cccc}
\hline
\multirow{2}{*}{Number of Intermediate Points} & \multicolumn{4}{c}{Cityscapes Dataset (\%)}                          \\ \cline{2-5} 
                                                &mIoU    &mF1     &mPre     &mRec     \\ \hline
\multirow{1}{*}{One Intermediate Point}         &50.49   &59.78   &64.72   &57.81    \\ 
\multirow{1}{*}{Two Intermediate Points}        &50.98   &60.15   &65.20   &58.25    \\
\multirow{1}{*}{Three Intermediate Points}      &\textbf{51.52}   &\textbf{60.76}   &65.36   &\textbf{59.02}    \\
\multirow{1}{*}{Four Intermediate Points}       &51.47   &60.73   &\textbf{65.44}   &58.81    \\
\multirow{1}{*}{Five Intermediate Points}       &50.49   &59.70   &65.00   &57.82    \\ \hline
\end{tabularx}
\label{tab:FedDSR_abl_number}
\vspace{-0.2cm}
\end{table}

In addition, \Cref{Fig:convergece_comparison} compares the convergence bewteen FedDSR and FedAvg using the Cityscapes dataset based on DeepLabv3+ model architecture. From \Cref{Fig:convergece_comparison}, we can conclude that the proposed FedDSR accelerates the convergence of FL training on both mIoU and mF1. For example, FedDSR-enabled setup reduces around (90-70) / 70 = 28.57\% in training rounds. This faster convergence of FedDSR helps to mitigate the computing pressure for computation resource-constrained vehicles in federated AD. 

In theory, the proposed FedDSR can be applied along with other existing FL algorithms synergistically. Therefore, we conduct comprehensive experiments on FedDSR with several FL algorithms (such as FedProx, FedDyn, FedGau, MOON, etc.) on the Cityscapes dataset using the DeepLabv3+ model architecture, and the results can be viewed in \Cref{tab:FedDSR_enhanced_FLs}. From this table, we can figure out following insights: (I) The proposed FedDSR can enhance the prediction performance of almost all existing FL algorithms across all adopted metrics. For example, FedDSR can enhance FedGau by (54.40-52.03) / 52.03 = 4.56\%, (64.43-62.23) / 62.23 = 3.54\%, (71.94-68.98) / 68.98 = 4.29\%, and (61.49-59.36) / 59.36 = 3.59\% in mIoU, mF1, mPre, and mRec, respectively. (II) For some FL algorithms, such as FedIR, SCAFFOLD, and BalanceFL, the proposed FedDSR may not improve their performance significantly. This is because they have integrated various forms of regularization or personalization, which may reduce the effectiveness of the FedDSR-incorporated intermediate supervision and regularization.

\begin{table}[tp]
\centering
\setlength{\tabcolsep}{4.5pt}
\caption{Quantitative performance comparison of FedDSR setups with different distances between adjacent points}
\begin{tabularx}{\linewidth}{c|cccc}
\hline
\multirow{2}{*}{Distance Between Intermediate Points} & \multicolumn{4}{c}{Cityscapes Dataset (\%)}                          \\ \cline{2-5} 
                                                &mIoU    &mF1     &mPre    &mRec     \\ \hline
\multirow{1}{*}{One-Base Distance}              &50.49   &59.70   &65.00   &57.82    \\ 
\multirow{1}{*}{Two-Base Distance}              &\textbf{51.72}   &\textbf{61.01}   &\textbf{65.54}   &\textbf{59.51}    \\
\multirow{1}{*}{Three-Base Distance}            &51.02   &60.27   &64.77   &58.39    \\ \hline
\end{tabularx}
\label{tab:FedDSR_abl_distance}
\vspace{-0.2cm}
\end{table}

\subsubsection{Qualitative Evaluation}
\Cref{tab:FedDSR_qualitive_comp} presents a qualitative performance comparison of FedDSR against FedAvg for multiple images based on the DeepLabv3+ model architecture. Based on \Cref{tab:FedDSR_qualitive_comp}, we can conclude that the proposed FedDSR achieves better prediction performance from detailed perspective. For example, for the last column of \Cref{tab:FedDSR_qualitive_comp}, FedDSR achieves the better segmentation between sky and road, relative to FedAvg. 

\subsection{Ablation Study} \label{sec:ablation}
This part reveals three types of ablation study: (I) how the number of intermediate points affects FedDSR's performance; (II) how the distance between adjacent intermediate points impacts FedDSR's performance; and (III) how positions of intermediate points affects FedDSR's performance.

\subsubsection{The Impact of the Number of Intermediate Points}
We compare five cases with different number of intermediate points ranging from 1 to 5, and the comparison results are illustrated in \Cref{tab:FedDSR_abl_number}, where we can conclude that cases with a moderate number of intermediate points perform best. This suggests no benefit in setting an excessive number of intermediate points in the proposed FedDSR method.

\subsubsection{The Impact of the Distance between Adjacent Intermediate Points}
We firstly define the base distance as a fixed number of layers between two adjacent layers. We subsequently conduct experiments by setting the distance between adjacent intermediate points as one base, two bases, and three bases. The experimental results are showcased in \Cref{tab:FedDSR_abl_distance}, which indicates that the case with two-base distance outclasses other two cases. This inspires us that in training a moderate distance facilitates a better performance.

\subsubsection{The Impact of the Position of Intermediate Points}
We position two intermediate points across three locations: Close-to-Input-Layer, Close-to-Central-Layer, and Close-to-Output-Layer. The experimental results are revealed in \Cref{tab:FedDSR_abl_position}. From \Cref{tab:FedDSR_abl_position}, we can figure out that the case of ``Close-to-Input-Layer'' consistently outperforms cases of ``Close-to-Central-Layer'' and ``Close-to-Output-Layer'', which inspires us that earlier supervision helps to enhance model performance larger. 

\begin{table}[tp]
\centering
\setlength{\tabcolsep}{6.8pt}
\caption{Quantitative performance comparison of FedDSR setups with different positions of intermediate points}
\begin{tabularx}{\linewidth}{c|cccc}
\hline
\multirow{2}{*}{Position of Intermediate Points} & \multicolumn{4}{c}{Cityscapes Dataset (\%)}                          \\ \cline{2-5} 
                                                &mIoU    &mF1     &mPre     &mRec     \\ \hline
\multirow{1}{*}{Close-to-Input-Layer}           &\textbf{52.98}   &\textbf{62.16}   &\textbf{67.21}   &\textbf{60.26}    \\ 
\multirow{1}{*}{Close-to-Central-Layer}         &51.95   &61.17   &65.68   &59.27    \\
\multirow{1}{*}{Close-to-Output-Layer}          &51.70   &60.97   &65.94   &59.36    \\ \hline
\end{tabularx}
\label{tab:FedDSR_abl_position}
\vspace{-0.2cm}
\end{table}

\section{Conclusion}
FL enables collaborative model training across distributed AD vehicles. However, FL faces challenge of poor generalization and slow convergence due to non-IID data. To address these issues, we propose FedDSR, a innovative framework incorporating multi-access intermediate layer supervision and negative entropy regularization. By supervising and penalizing the learning of feature representations at intermediate stages, FedDSR improves model generalization and accelerates convergence in AD scenarios. Evaluated on AD semantic segmentation task across various model architectures and existing FL algorithms, FedDSR achieves better performance and faster convergence, demonstrating its suitability and practicality for federated AD ecosystems.

\appendix
This appendix demonstrates the primary proof sketch of \textit{Theorem 1} as following steps:
\begin{enumerate}[leftmargin=*]
\item \textbf{Vehicle Gradient Decomposition:}
\begin{align}
\!\!\!g_n\!\!=&\nabla \mathcal{L}^{n}(\theta) = \nabla \mathcal{L}_{\text{CE}}^{n}(\theta) + \nonumber \\ &\sum\nolimits_{m=1}^M \!(\alpha_m \nabla \mathcal{L}_{\text{MI}}^{m,n}(\theta,\phi_m) \!\!+\!\! \lambda_m \nabla \mathcal{L}_{\text{NE}}^{m,n}(\theta)).
\end{align}
Global gradient is $\nabla \mathcal{L}(\theta) = \sum_{n=1}^N w_n \nabla \mathcal{L}^{n}(\theta)$.

\item \textbf{Local Descent with L-smoothness:}
For each component $s$ and client $n$, using L-smoothness of $\mathcal{L}_s^{n}$ and a local step $\theta_{n}^{t,\ell+1} = \theta_{n}^{t,\ell} - \eta_t g_{n}^{t,\ell}$ (stochastic gradient),
\begin{align}
\mathcal{L}_s^{n}(\theta_{n}^{t,\ell+1}) \le &\mathcal{L}_s^{n}(\theta_{n}^{t,\ell}) - \eta_t \langle \nabla \mathcal{L}_s^{n}(\theta_{n}^{t,\ell}), g_{n}^{t,\ell} \rangle + \nonumber \\ &(\mathcal{L}_s^{n} \eta_t^2) / 2 \|g_{n}^{t,\ell}\|^2.
\end{align}
Summing over $s$ yields an analogous inequality for $\mathcal{L}^{n}$.

\item \textbf{Variance Bounds and Mini-batch Noise:}
Taking expectations over the local minibatch randomness and applying bounded gradient/variance assumptions give
\begin{align}
\mathbb{E}[\mathcal{L}^{n}(\theta_{n}^{t,\ell+1})] \le &\mathbb{E}[\mathcal{L}^{n}(\theta_{n}^{t,\ell})] - \eta_t \mathbb{E}\|\nabla \mathcal{L}^{n}(\theta_{n}^{t,\ell})\|^2 + \nonumber \\ &\frac{\mathcal{L}_{\text{max}} \eta_t^2}{2} ((G_{\text{T}}^{n})^2 + (\sigma_{\text{T}}^{n})^2),
\end{align}
with $G_{\text{T}}^{n2}$, $\sigma_{\text{T}}^{n2}$ defined analogously per vehicle.

\item \textbf{Accumulating $E$ Local Steps and Federated Aggregation:}
After $E$ steps, each client produces $\theta_{n}^{t,E}$. The server forms
$\theta_{t+1} = \sum_{n \in S^t} w_n^t \, \theta_{n}^{t,E}$.
The difference between $\theta_{t+1}$ and $\theta_t$ introduces a \emph{drift} term that can be bounded using smoothness and the heterogeneity measure. 
\begin{align}
\!\!&\!\!\mathbb{E}[\mathcal{L}(\theta_{t+1})] \le \mathbb{E}[\mathcal{L}(\theta_t)] - \frac{\eta_t}{2}\mathbb{E}\|\nabla \mathcal{L}(\theta_t)\|^2 + \nonumber \\ \!\!\!\!&\frac{\mathcal{L}_{\text{max}}\eta_t^2}{2}(G_{\text{T}}^2 \!+\! \sigma_{\text{T}}^2) \!+\! c \mathcal{L}_{\text{max}} \eta_t^2 E^2 (\mathbb{E}\|\nabla \mathcal{L}(\theta_t)\|^2 \!\!+\!\! \mathcal{H}),
\end{align}
where constant $c>0$ depends on participation and smoothness, the heterogeneity measure $\mathcal{H}$ is defined as $\sum\nolimits_{n=1}^N w_n \, \|\nabla \mathcal{L}^{n}(\theta_t) - \nabla \mathcal{L}(\theta_t)\|^2$.

\item \textbf{Summation over $t=1$ to $T$ and Telescoping:}
Summing and rearranging based on $\eta_t=\eta/\sqrt{T}$, we obtain
\begin{align}
\frac{1}{T}\sum_{t=1}^T \mathbb{E}\|\nabla \mathcal{L}(\theta_t)\|^2 \le &\frac{2\Delta}{\eta \sqrt{T}} + \frac{\mathcal{L}_{\text{max}} \eta}{\sqrt{T}}\left(G_{\text{T}}^2 + \sigma_{\text{T}}^2\right) + \nonumber \\ &\frac{\mathcal{L}_{\text{max}} \eta}{\sqrt{T}} \, \Gamma_{\text{drift}},
\end{align}
where $\Gamma_{\text{drift}} \le c E^2 (\frac{1}{T}\sum_{t=1}^T \mathbb{E}\|\nabla \mathcal{L}(\theta_t)\|^2 + \mathcal{H})$ is absorbed into the RHS. With sufficiently small $\eta E$, the drift contribution is controlled, yielding Theorem~\ref{thm:federated_convergence}.
\end{enumerate}

\end{document}